\documentclass[letterpaper]{article} 
\usepackage{aaai23}  
\usepackage{times}  
\usepackage{helvet}  
\usepackage{courier}  
\usepackage[hyphens]{url}  
\usepackage{graphicx} 
\usepackage{natbib}  
\usepackage{caption} 
\usepackage{algorithm}
\usepackage{algorithmic}

\usepackage{newfloat}
\usepackage{listings}
\DeclareCaptionStyle{ruled}{labelfont=normalfont,labelsep=colon,strut=off} 
\floatstyle{ruled}
\newfloat{listing}{tb}{lst}{}
\floatname{listing}{Listing}
\usepackage{multirow}
\usepackage{subcaption}

\usepackage{xcolor}
\definecolor{uni-yellow}{HTML}{EFC400}
\definecolor{bi-orange}{HTML}{FF8900}
\definecolor{multi-blue}{HTML}{000CC7}
\newcommand{\mc}[1]{\multicolumn{1}{c}{#1}}

\usepackage{amsmath}
\usepackage{amssymb}
\usepackage{amsfonts}       
\usepackage{comment}
\DeclareMathOperator*{\argmin}{arg\,min}
\usepackage[symbol]{footmisc}
\usepackage{booktabs}       

\title{AutoFraudNet: A Multimodal Network to Detect Fraud in the Auto Insurance Industry}
\author{
    Azin Asgarian \textsuperscript{\rm \dag} \textsuperscript{\rm 1},
    Rohit Saha \textsuperscript{\rm \dag} \textsuperscript{\rm 1},
    Daniel Jakubovitz \textsuperscript{\rm \dag} \textsuperscript{\rm 2},
    Julia Peyre \textsuperscript{\rm \dag} \textsuperscript{\rm 2},
}
\affiliations {
    azin@georgian.io, rohit.saha@georgian.io, daniel.jakubovitz@tractable.ai, julia.peyre@gmail.com\\
    \textsuperscript{\rm 1} Georgian.io, Toronto, Canada\\
    \textsuperscript{\rm 2} Tractable.ai, London, United Kingdom \\
}

\begin{document}
\maketitle

\begin{abstract}
\label{sec:abstract}
In the insurance industry detecting fraudulent claims is a critical task with a significant financial impact.
A common strategy to identify fraudulent claims is looking for inconsistencies in the supporting evidence. However, this is a laborious and cognitively heavy task for human experts as insurance claims typically come with a plethora of data from different modalities (e.g. images, text and metadata). To overcome this challenge, the research community has focused on multimodal machine learning frameworks that can efficiently reason through multiple data sources.
Despite recent advances in multimodal learning, these frameworks still suffer from (i) challenges of joint-training caused by the different characteristics of different modalities and (ii) overfitting tendencies due to high model complexity.
In this work, we address these challenges by introducing a multimodal reasoning framework, \emph{AutoFraudNet}\footnote{Automobile Insurance Fraud Detection Network}, for detecting fraudulent auto-insurance claims. AutoFraudNet utilizes a cascaded slow fusion framework and state-of-the-art fusion block, BLOCK Tucker, to alleviate the challenges of joint-training. Furthermore, it incorporates a light-weight architectural design along with additional losses to prevent overfitting.
Through extensive experiments conducted on a real-world dataset, we demonstrate: (i) the merits of multimodal approaches, when compared to unimodal and bimodal methods, and (ii) the effectiveness of AutoFraudNet in fusing various modalities to boost performance (over 3\% in PR AUC).
\end{abstract}

\section{Introduction}
\label{sec:introduction}
The insurance industry is susceptible to fraud attempts, namely customers who try to obtain illicit financial benefits using untruthful claims. Specifically, the auto-insurance industry incurs financial losses of at least \$29 billion globally every year due to fraud~\cite{insurance_loss}. Moreover, almost 30\% of submitted auto-insurance claims contain fraudulent elements but only less than 3\% of such claims are prosecuted~\cite{nian2016unsupervised}.
Fraudulent insurance claims can generally be identified based on inconsistencies in the supporting evidence~\cite{dionne2005humanexpert}.
However, given the huge volume of data that is typically processed in the insurance industry, manual inspection is costly, time-consuming, and prone to mistakes~\cite{artis2002manualinspectiontakestime,bolton2002hugevolumes}.
These limitations, combined with the need for automation, have given rise to machine learning (ML) research in the field of fraud detection.

\vspace{.1cm}
ML techniques have been adopted to detect fraudulent claims across several insurance industries; from healthcare~\cite{rawte2015healthcarefraud} to real-estate~\cite{severino2021propertyFraud} to automobile~\cite{pathak2005fuzzy, viaene2002sotaFraud, brockett2002risk}. 
In the context of the auto-insurance industry, previous works have explored both supervised~\cite{fraud_detection_research, harjai2019smote} and unsupervised~\cite{nian2016unsupervised} approaches.
While previous studies mainly focus on tabular data, some recent works have explored other modalities such as textual~\cite{LDA_based_DL_for_automobile_insurance_fraud_detection} and visual~\cite{li2018yoloFraud} data.
As a major drawback, the proposed methods are all limited to a \emph{single} modality and cannot leverage the complementary information offered by \emph{multiple} modalities. Considering the multimodal nature of claims in the auto-insurance industry, this can lead to inaccurate assessments and hence loss of client trust.

\vspace{.1cm}
Multimodal reasoning therefore lends itself as a natural solution to this problem. Recently, several prominent works~\cite{clip, Li2019VisualBERTAS, Wang2021SimVLMSV, dou2022an} have emerged with impressive results on visual-language reasoning tasks. In parallel, ~\cite{Lahat2015MultimodalDF, Data_fusion_an_overview, Deep_Multimodal_Fusion} explicitly investigate efficient strategies to fuse different data modalities.
Despite these advances, applications of multimodal frameworks in the insurance industry are fairly limited, mostly due to two main challenges~\cite{wang2020makes}.
First, different modalities overfit and generalize at different rates which makes joint-training difficult. Second, the inherent increased capacity of these models makes them prone to overfitting, which is further exacerbated by data quality issues (e.g. data scarcity, class imbalance, prevalence of noise, etc.).
While there have been several attempts to overcome these limitations, the proposed solutions
fall short when applied in real-world settings~\cite{wang2020makes}.

\maketitle
\def\thefootnote{\dag}\footnotetext{All authors have contributed equally to this work.}

\paragraph{Contributions.}
In this paper, we propose a multimodal reasoning framework, AutoFraudNet, that addresses the aforementioned challenges. To the best of our knowledge, this is the first attempt to use multimodal reasoning for the detection of fraudulent auto-insurance claims.
Our framework leverages three modalities that are commonly present in auto-insurance claims: images, text, and tabular data. To effectively fuse these modalities together, our approach uses a cascaded slow fusion framework and state-of-the art fusion blocks. To combat overfitting, we use a lightweight design with more granular supervision provided by supplementary loss functions. By conducting experiments on a real-world dataset, we show the effectiveness of our framework. We also demonstrate the merits of moving from a single data modality to bimodal and multiple modality frameworks.

\section{Related work}
\label{sec:related_work}
Previous works have applied a wide range of data mining~\cite{fraud_detection_research}, unsupervised ~\cite{nian2016unsupervised}, and supervised ML techniques~\cite{prasasti2020supervised} to the problem of fraud detection.
These techniques entail training light-weight classifiers, including Random Forests~\cite{itri2019comparativeML}, SVMs~\cite{muranda2020svm} and Naïve  Bayes~\cite{detecting_fraud_in_insurance_claims}.
As class-imbalance is prevalent in auto-insurance datasets, intelligent sampling techniques such as SMOTE~\cite{harjai2019smote} and ADASYN~\cite{muranda2020svm} have been explored to improve model generalization. Moving beyond tabular data and classical machine learning methods,~\cite{LDA_based_DL_for_automobile_insurance_fraud_detection} trains a deep neural network with features extracted by Latent Dirichlet allocation from textual descriptions. ~\cite{li2018yoloFraud} investigates the visual modality by training a YOLO model~\cite{redmon2015yolo} to detect whether damage is present in images of vehicles. The above methods operate with a \emph{single} data modality, disregarding the abundance of complementary information that is available in other modalities.

Recently, several works~\cite{dou2022an} introduced multimodal networks that combine information across different data modalities. Notably, CLIP~\cite{clip} leverages natural language supervision to learn generalizable visual concepts and shows impressive zero-shot capabilities.
Much of the success of these works can be attributed to the underlying transformer architecture~\cite{vaswani2017attention}, and the abundance of paired image-text data. 

However, training these high-capacity models end-to-end in settings with limited data and compute power is very challenging. To combat these issues, recent works \cite{GaleNet} use pre-extracted features instead of the raw data, shifting the focus more toward the fusion mechanism. To combine features extracted from pre-trained networks~\cite{atito2021sit, qiu2020plm}, various fusion strategies have emerged~\cite{Lahat2015MultimodalDF, a_survey_on_data_fusion, Deep_Multimodal_Fusion}.
Some studies explore the optimal stage at which fusion should occur, i.e. early~\cite{barnum2020earlyfusion}, intermediate~\cite{roitberg2019intermediatefusion} or late ~\cite{zhang2019latefusion}, whereas others ~\cite{BLOCK_fusion, kim2016hadamard} explore the design of fusion operations.

Our work leverages pre-trained models as feature extractors for images and text, and categorical representations of the tabular data. The extracted features are combined via a slow fusion paradigm to facilitate cross-modal interactions.

\section{Methodology}
\label{sec:methodology}

\subsection{Problem formulation}
\label{subsec:problem_formulation}
We formulate the problem as a binary classification task. Given the visual, textual and tabular data submitted with an auto insurance claim, the goal is to predict whether the claim is fraudulent or not.
The visual evidence comes in the form of multiple images of the damaged vehicle. The textual and tabular information, however, contain high-level contextual and circumstantial information about the claim.

\subsection{Dataset \& Model} 
\label{subsec:dataset_and_model}
Our multimodal dataset contains Japanese auto insurance claims characterized by various modalities. Each claim is represented by a unique \textit{id}, which we use as an anchor point to align all modalities: (i) visual, (ii) textual and (iii) tabular.
A claim can address multiple vehicle parts (e.g. front left door and front bumper), where typically a vehicle is partitioned to 21 main parts.
Since our proposed framework relies on the availability of all features, we filter out claims that do not meet this condition. Following this selection criteria, there remain one million claims in our dataset with 30,000 (3\%) belonging to the fraudulent class. To run experiments, we split this data into training, validation and test sets using a 80\%-10\%-10\% ratio and stratified sampling.

\subsubsection{Visual Data}
\label{subsec:visual_data}
As mentioned in section~\ref{subsec:dataset_and_model}, claims contain visual data in the form of multiple images. To extract visual features, each image in the claim is passed through two in-house pre-trained CNNs\footnote{EfficientNet based \cite{pmlr-v97-tan19a-efficientnet}.}: CDS (crack/dent/scratch) and UD (undamaged/damaged). Given an image of a vehicle, the UD network detects the presence of damage and its severity on the vehicle, and the CDS network classifies the type of damage.
As these networks were pre-trained on large volumes of similar visual data, they lend themselves as relevant feature extractors.
Next, the extracted image-level CDS and UD features are fed to additional CDS and UD encoder blocks to obtain claim-level aggregated features. This process is summarized as follows:
Let $\{I_1, I_2, ..., I_n\}$ represent the set of images available in a claim. Each image is fed to the CDS and UD networks, resulting in two 720-dimensional embeddings. Let $\{E^{cds}_{1}, E^{cds}_{2}, ..., E^{cds}_{n}\} \in \mathbb{R}^{n \times 720}$ and $\{E^{ud}_{1}, E^{ud}_{2}, ..., E^{ud}_{n}\} \in \mathbb{R}^{n \times 720}$ be the sets of image-level features corresponding to the two networks. Each feature-set is then passed through a respective MLP encoder block to obtain claim-level features: $\{A_{\text{CDS}}, A_{\text{UD}}\} \in \mathbb{R}^{50}$. These encoder blocks are trained as part of our framework to (i) learn intermediate latent representations corresponding to each image-level feature via two fully connected layers, and (ii) aggregate the latent representations to a claim level representation via average pooling.

\subsubsection{Textual Data}
\label{subsec:textual_data}
The textual data provides a description of the circumstances of the claim in Japanese.
We make use of a pre-trained canonical (English) BERT model~\cite{devlin2018bert}, and fine-tune it on the Japanese descriptions using the Masked-Language-Modeling (MLM) objective.
Using this model, a claim's textual description can now be represented by a 768-dimensional embedding: $A_{\text{Text}}$.

\vspace{.1cm}
\subsubsection{Tabular Data}
\label{subsec:tabular_data}
\paragraph{Structural Features.} These are categorical features relating to metadata submitted with the claim, e.g. whether the vehicle was in movement when the accident occurred, the accident's point of impact, etc. Each categorical feature is represented in the form of a 1-hot encoding, and does not undergo any additional pre-processing. These features are represented by an 87-dimensional embedding: $A_{\text{Struct}}$.

\paragraph{Visibility Scores.} Similar to visual features, these scores are extracted from two in-house pre-trained CNNs: (i) the aforementioned UD (undamaged/damaged) and (ii) Part-Visibility.
The Part-Visibility network detects the presence of the 21 main vehicle parts (e.g. the back bumper) in the claim images.
Unlike the visual features, in this case we use the networks' post-softmax output scores rather than the visual embeddings.
The extracted scores from both networks are $21$ dimensional vectors relating to the $21$ main vehicle parts. When a part is absent in the image, its Part-Visibility and UD scores are respectively inferred as invisible and undamaged. Let $\{A_{Part}, A_{UD;Tab}\} \in \mathbb{R}^{n \times 21}$ denote the 2 sets of scores corresponding to $n$ images in a given claim. To obtain claim-level scores, the image-level scores for each claim are aggregated by computing the max, min and average, yielding $\{A_{Part}, A_{UD;Tab}\} \in \mathbb{R}^{3 \times 21}$, where $3$ represents the number of statistical operations. Finally, $\{A_{Part}, A_{UD;Tab}\}$ are unrolled and concatenated, forming the claim-level feature $A_{\text{SPUD}} \in \mathbb{R}^{126}$.

\begin{figure*}[tbh] 
\includegraphics[width=.65\paperwidth]{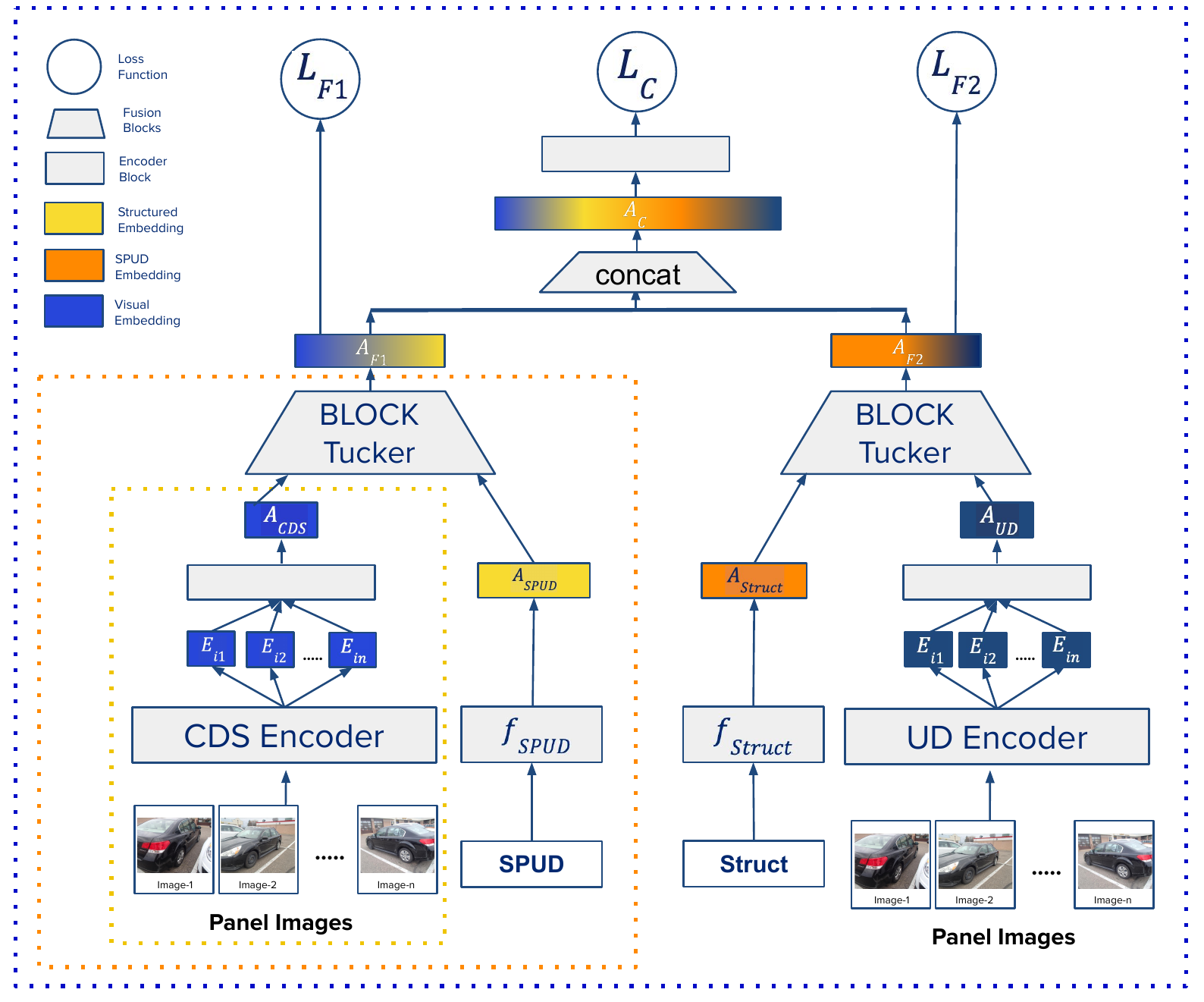}
\centering
\caption{An overview of our proposed framework. The \textcolor{uni-yellow}{yellow}, \textcolor{bi-orange}{orange} and \textcolor{multi-blue}{blue} dotted boxes contain the \textcolor{uni-yellow}{uni-}, \textcolor{bi-orange}{bi-} and \textcolor{multi-blue}{multi}-modal components respectively. In \textit{AutoFraudNet}, we use $L_C$ as the final loss for optimization. In \textit{AutoFraudNet + Heads}, however, we employ the two losses $L_{F1}$ and $L_{F2}$ in addition to $L_C$ for training. Our framework excludes the usage of the textual modality owing to its inferior performance on the task at hand.}
\label{fig:architecture}
\end{figure*}

\section{Experiments}
\label{sec:experiments}
To assess the contribution of each data modality and the benefits of fusing them together, we conduct three sets of experiments.
First, we begin with unimodal experiments, where we evaluate and compare the features of each modality separately (section~\ref{subsec:exp-unimodal}).
Second, we conduct bimodal experiments where we ablate over pairs of features and fusion strategies (section~\ref{subsec:exp-bimodal}).
Finally, we conduct multimodal experiments using the most performative feature pairs and fusion strategies from the previous setups (section~\ref{subsec:exp-multimodal}).
To ensure a fair comparison, we use the same training and evaluation settings across all experiments, as discussed in section~\ref{subsec:exp-metrics}.

\subsection{Unimodal}
\label{subsec:exp-unimodal}
To evaluate and compare different modalities and their respective features, we process each feature using a neural network of two fully-connected layers. Both layers contain 500 neurons and use the ReLU~\cite{nair2010rectified} activation, followed by dropout~\cite{srivastava2014dropout} with $p=0.5$. The output of the network is then fed into a softmax layer for classification.

\subsection{Bimodal}
\label{subsec:exp-bimodal}
To assess the benefits of cross-modal interactions for our task, we extend the unimodal setup to conduct bimodal experiments. We begin by selecting feature pairs such that each feature represents a different modality.
For instance, feature CDS from the visual modality is paired with feature Text from the textual modality. Following this selection criteria, $8$ cross-modal feature pairs are created. To facilitate bimodal reasoning, a pair of features is fed into a fusion module. Finally, the output of the fusion module is passed into a softmax layer for classification.

Following the success of popular fusion strategies, we conduct extensive bimodal experiments using the canonical versions of (i) Concat MLP~\cite{BLOCK_fusion}, (ii) Linear Sum~\cite{BLOCK_fusion}, (iii) BLOCK~\cite{BLOCK_fusion}, (iv) BLOCK Tucker~\cite{BLOCK_fusion}, (v) MLB~\cite{kim2016hadamard}, (vi) MFH~\cite{yu2018beyond}, and (vii) MFB~\cite{yu2017multi}.

\subsection{Multimodal}
\label{subsec:exp-multimodal}
We show the effectiveness of our proposed multimodal method by benchmarking it against several other approaches.

\paragraph{Concat MLP.} We begin by concatenating all features from each modality. Let $A_{all}=[A_{\text{CDS}}, A_{\text{UD}}, A_{\text{SPUD}}, A_{\text{Struct}}, A_{\text{Text}}]$ be the concatenated representation. $A_{all}$ is then passed through a network similar to the setup described in section~\ref{subsec:exp-unimodal}. We name this configuration \mbox{\textit{Concat MLP - All}}. Additionally, we train a variation of the network that concatenates all features except Text. Due to the limited amount of textual Japanese descriptions available for training, we posit that the textual features are not refined enough to boost model performance through the fusion process.
We name this variation \mbox{\textit{Concat MLP - w/o Text}}.

\paragraph{Slow Fusion.} We observe that an early-stage naïve concatenation of features, extracted from different modalities, can lead to challenges in training~\cite{wang2020makes}. To combat these problems, we use \textit{slow-fusion} (SF), a common paradigm in which features from various modalities are gradually fused together in a cascaded manner~\cite{Joze_2020_CVPR}.

To this end, the multimodal framework jointly trains two bimodal modules whose intermediate activations $(A_{F1}, A_{F2})$ are then fed into an additional fusion layer, as depicted in Figure~\ref{fig:architecture}. Since the number of possible configurations to explore in this setup is extremely high, we leverage our findings from the bimodal experiments to more efficiently navigate through the search space.
More concretely, we fix the fusion strategy in the first layer of our SF framework to the one that obtains the best bimodal results.
For the second fusion layer, however, we experiment with the top four best-performing fusion strategies from the bimodal setup.
We name the resultant configurations \textit{SF - $<$fusion strategy$>$}.

\paragraph{Our framework.}
As noted in section~\ref{sec:introduction}, the task at hand has several challenges, including scarcity of data and severe class imbalance. While fusion strategies are effective at learning cross-modal interactions, they do come with a significant price: millions of learnable parameters. Therefore, they should be judiciously used, failure of which can lead to overfitting issues~\cite{wang2020makes}.
With this notion in mind, our proposed framework (Figure~\ref{fig:architecture}) uses the SF framework with a single-layer neural network for the second fusion step, i.e. the intermediate representations of the two fusion blocks are concatenated, $A_C = [A_{F1}, A_{F2}]$, passed through a single fully connected layer and then fed into a softmax layer for classification. We name this framework \emph{AutoFraudNet}.

Inspired by the idea behind the inception network~\cite{lin2013network, szegedy2015going}, we introduce a further improvement: each of ${A_{F1}, A_{F2}}$ is additionally fed into a separate classification head to provide more granular supervision during training. We name this variation \emph{AutoFraudNet + Heads}.

Combining all the losses computed from the three classification layers, the overall optimization objective becomes:
\begin{equation}
    L = \underset{\theta}{\argmin} \{ L_{F1} + L_{F2} + L_{C} \},
\end{equation}
where $L_{F1}$ and $L_{F2}$ correspond to the losses computed from $A_{F1}$ and $A_{F2}$, $L_{C}$ corresponds to the loss computed from $A_{C}$, and $\theta$ is the set of learnable parameters.

\subsection{Settings}
\paragraph{Training Configuration}
In all cases (uni-, bi- and multi-modal experiments) we employ the Cross-Entropy loss. We use the Adam optimizer \cite{kingma2014adam}, initialized with a learning rate of $1e{-}3$. To prevent overfitting, we use Early Stopping \cite{caruana2000overfitting} with the patience set to 3.
To counter the class imbalance, we sample data in a balanced manner such that every mini-batch consists of 50\% samples from each class.
Finally, we run each model $5$ times with random initialization of weights and report the averaged metrics along with the respective standard deviation values.

\paragraph{Evaluation Metrics}
\label{subsec:exp-metrics}
To compare the different experimental settings, we measure the Precision (P), Recall (R), and F1-Score (F1) for both the Fraudulent and Not Fraudulent classes.
We tune the decision threshold of each model to reach at least 80\% recall on the Fraudulent class to satisfy business requirements.
Additionally, we use two other metrics to obtain a more holistic view of model performance: (i) Area Under the Precision-Recall Curve (PR AUC) which is a threshold independent metric and (ii) Balanced Accuracy (Bal. Acc.) which is the unweighted average of recall values obtained on each class, commonly used in imbalanced classification scenarios. We use PR AUC as our primary metric for comparison since unlike the other metrics, it is threshold independent and provides a more holistic view of model performance.

\section{Results}
\label{sec:results}
In this section we present our findings from the experiments described in section~\ref{sec:experiments}.
We first discuss the results from the uni-, bi- and multi-modal setups respectively in Sections~\ref{subsec:res-unimodal}, \ref{subsec:res-bimodal} and \ref{subsec:res-multimodal}.
We then share our insights by comparing all three setups in section~\ref{subsec:res-insights}.

\subsection{Unimodal}
\label{subsec:res-unimodal} 
Table~\ref{table:res-unimodal} shows the individual performance of each feature. In general, the visual features are the most performative followed by tabular and textual features. The CDS and UD embeddings achieve the highest PR AUC scores, highlighting the importance of visual features in detecting fraudulent claims.
The performance of the textual and structural features is lower compared to visual features. This is most likely due to lower level of relevance and granularity of the information embodied in them.
Feature SPUD performs better on the task than Struct and Text features. We reason that this is because the Struct and Text features contain high-level information, which just by itself is not discriminative enough.
In contrast, the SPUD feature is very performative as it is a highly refined low-level visual-proxy.

\begin{table*}[ht]
\centering
\caption{Performance comparison of various unimodal features}
\begin{tabular}{rccccccccc} 
\toprule
\multirow{2}{*}{Features} & \multirow{2}{*}{Modality}& \multirow{2}{*}{PR AUC} & \multirow{2}{*}{Bal. Acc.} & \multicolumn{3}{c}{Fraudulent} & \multicolumn{3}{c}{Not Fraudulent}\\
\cmidrule(lr){5-7} \cmidrule(l){8-10}
& & & & \mc{P} & \mc{R} & \mc{F1} & \mc{P} & \mc{R} & \mc{F1}\\
\midrule
CDS             & Visual & 0.194 & 0.755 & 0.094 & 0.811 & 0.168 & 0.989 & 0.699 & 0.819 \\
UD              & Visual & 0.183 & 0.547 & 0.074 & 0.810 & 0.136 & 0.988 & 0.611 & 0.755 \\
SPUD            & Tabular & 0.160 & 0.710 & 0.043 & 0.811 & 0.083 & 0.977 & 0.317 & 0.479 \\
Struct          & Tabular & 0.065 & 0.564 & 0.043 & 0.810 & 0.083 & 0.977 & 0.317 & 0.478 \\
Text            & Textual & 0.060 & 0.563 & 0.046 & 0.962 & 0.088 & 0.197 & 0.133 & 0.159 \\
\bottomrule
\end{tabular}
\label{table:res-unimodal}
\end{table*}

\subsection{Bimodal}
\label{subsec:res-bimodal}
\begin{figure*}[ht]
    \centering
    \begin{subfigure}[t]{0.5\textwidth}
        \centering
        \includegraphics[width=\textwidth]{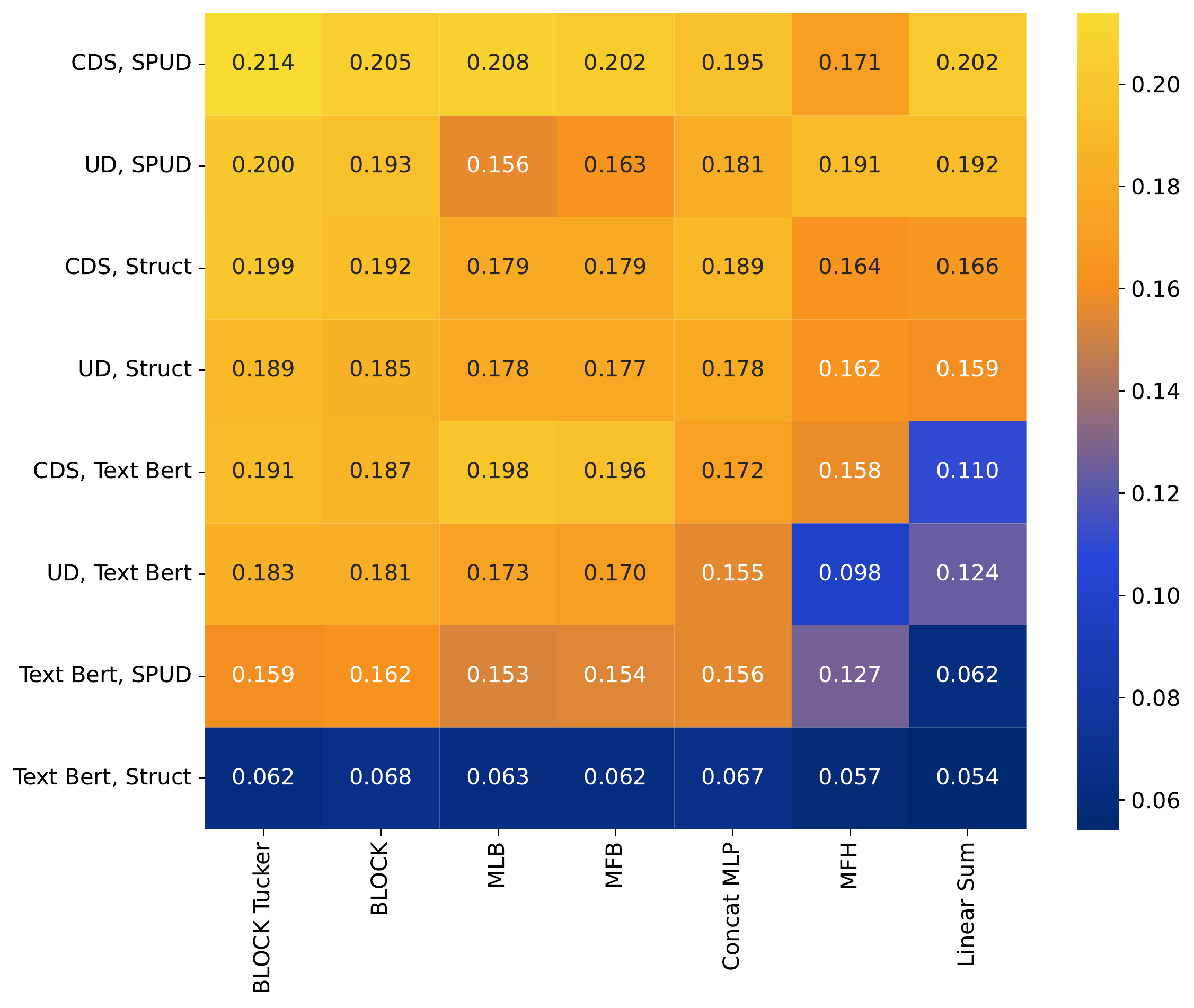}
        \caption{Average of PR AUC}
        \label{subfig:bi_heatmap_avg}
    \end{subfigure}%
    ~~ 
    \begin{subfigure}[t]{0.5\textwidth}
        \centering
        \includegraphics[width=\textwidth]{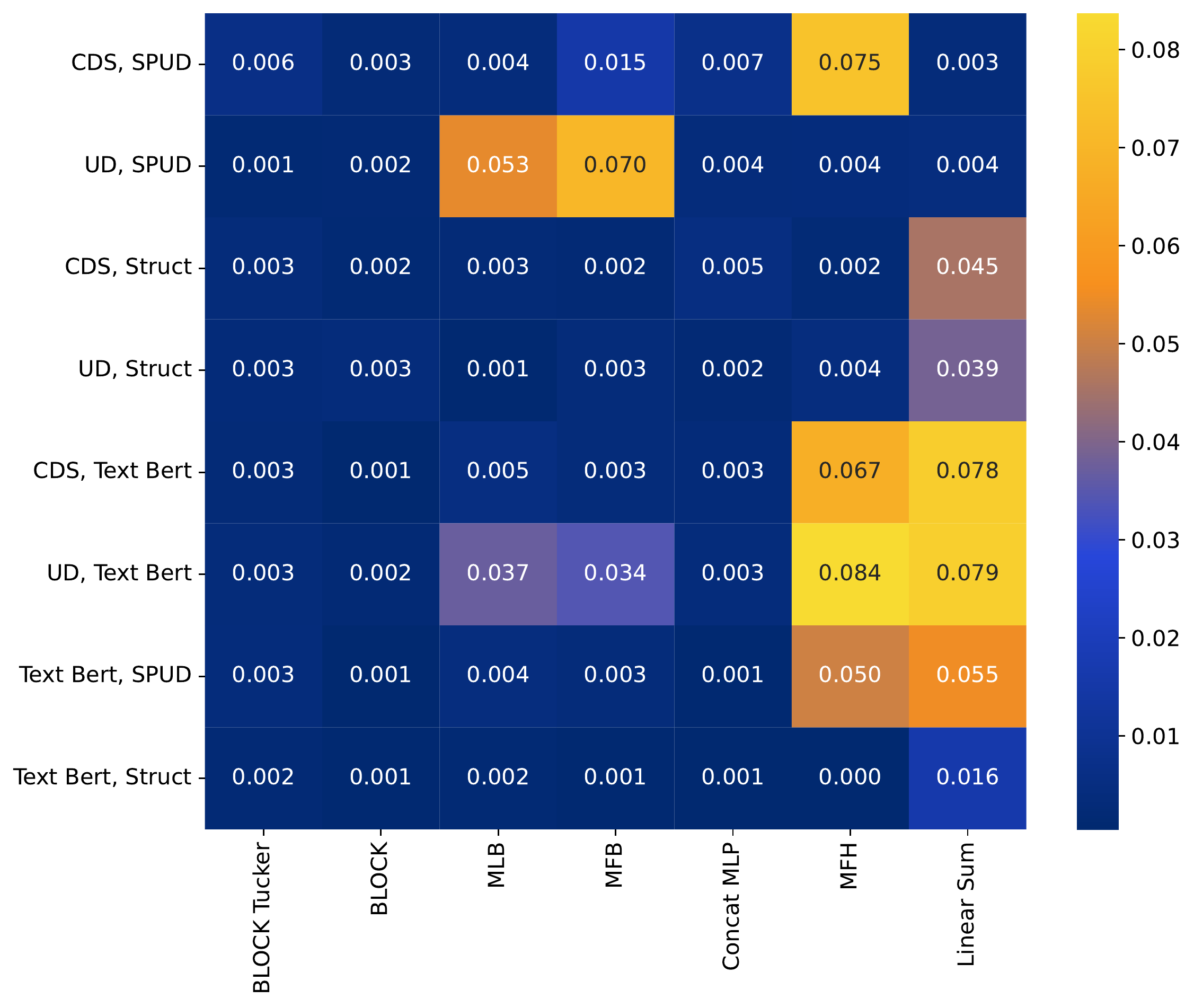}
        \caption{Standard Deviation of PR AUC}
        \label{subfig:bi_heatmap_std}
    \end{subfigure}
    \caption{Comparison of PR AUC across all combinations of feature pairs and fusion strategies. The values are obtained by computing the (a) average and (b) standard deviation of PR AUC for each setting across five runs with randomly initialized weights.}
    \label{fig:bi_heatmap}
\end{figure*}

In this section, we discuss the performance of $7$ fusion strategies and $8$ feature pairs. The average and standard deviation of PR AUC for all $56$ possible combinations of feature pairs and fusion strategies is presented in Figures~\ref{subfig:bi_heatmap_avg} and \ref{subfig:bi_heatmap_std} respectively. 

From these figures, we can conclude that the visual features fused with tabular features yield strong performance. In particular, the feature pairs (CDS, SPUD) and (UD, Struct) obtain the best results. This implies that there is cross-modal interaction between visual features and tabular features, each providing complementary information for identifying fraudulent claims. We also observe that the Text feature, in conjunction with other modalities, does not perform well. This further corroborates our finding from section~\ref{subsec:res-unimodal} where Text obtains the lowest unimodal performance.

\begin{figure*}
\centering
\begin{minipage}{.48\textwidth}
  \centering
  \includegraphics[width=1\linewidth]{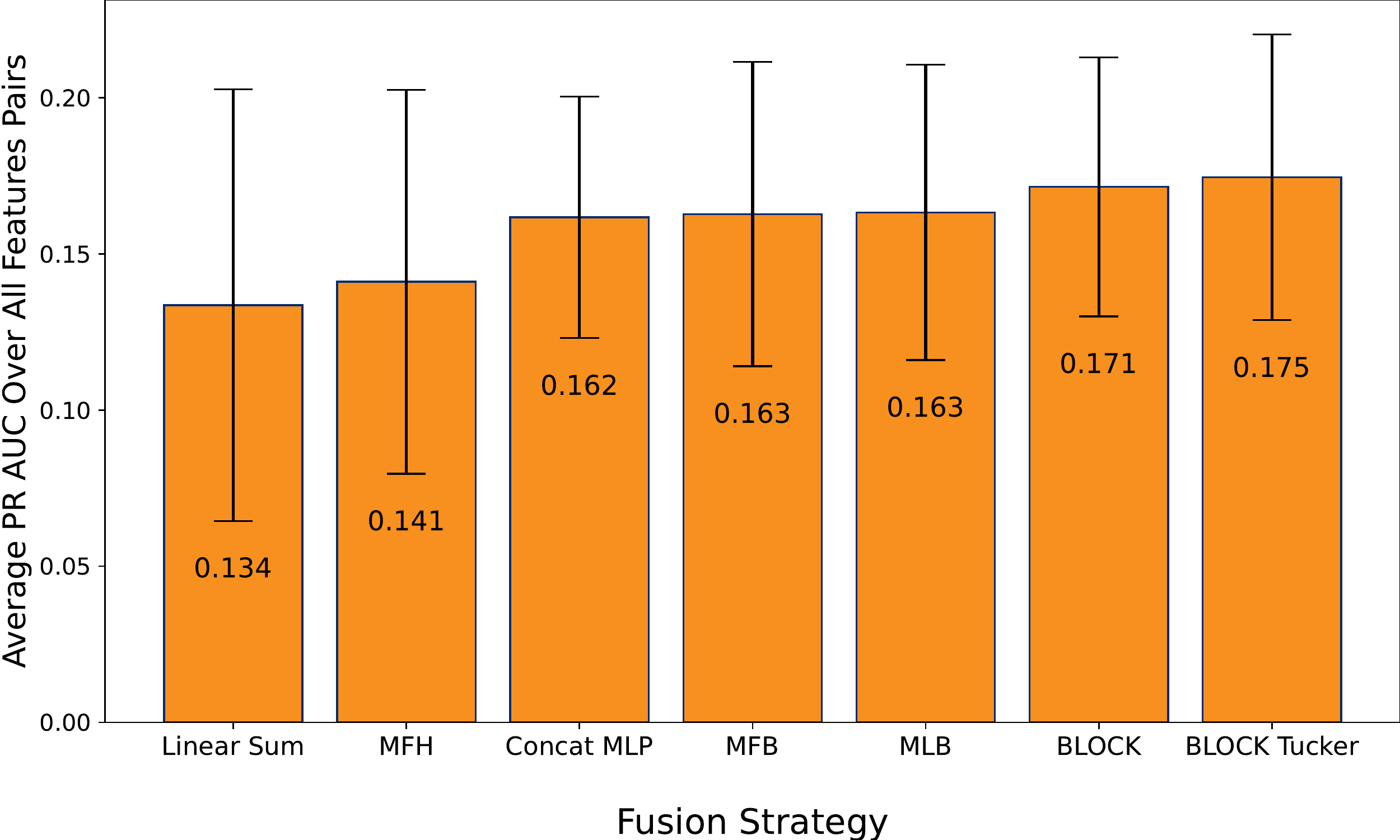}
  \captionof{figure}{Average and standard deviation of PR AUC of the bimodal fusion strategies. The corresponding average PR AUC is shown on each bar.}
  \label{fig:bi_FBs}
\end{minipage}%
\hfill
\begin{minipage}{.48\textwidth}
  \centering
  \includegraphics[width=1\linewidth]{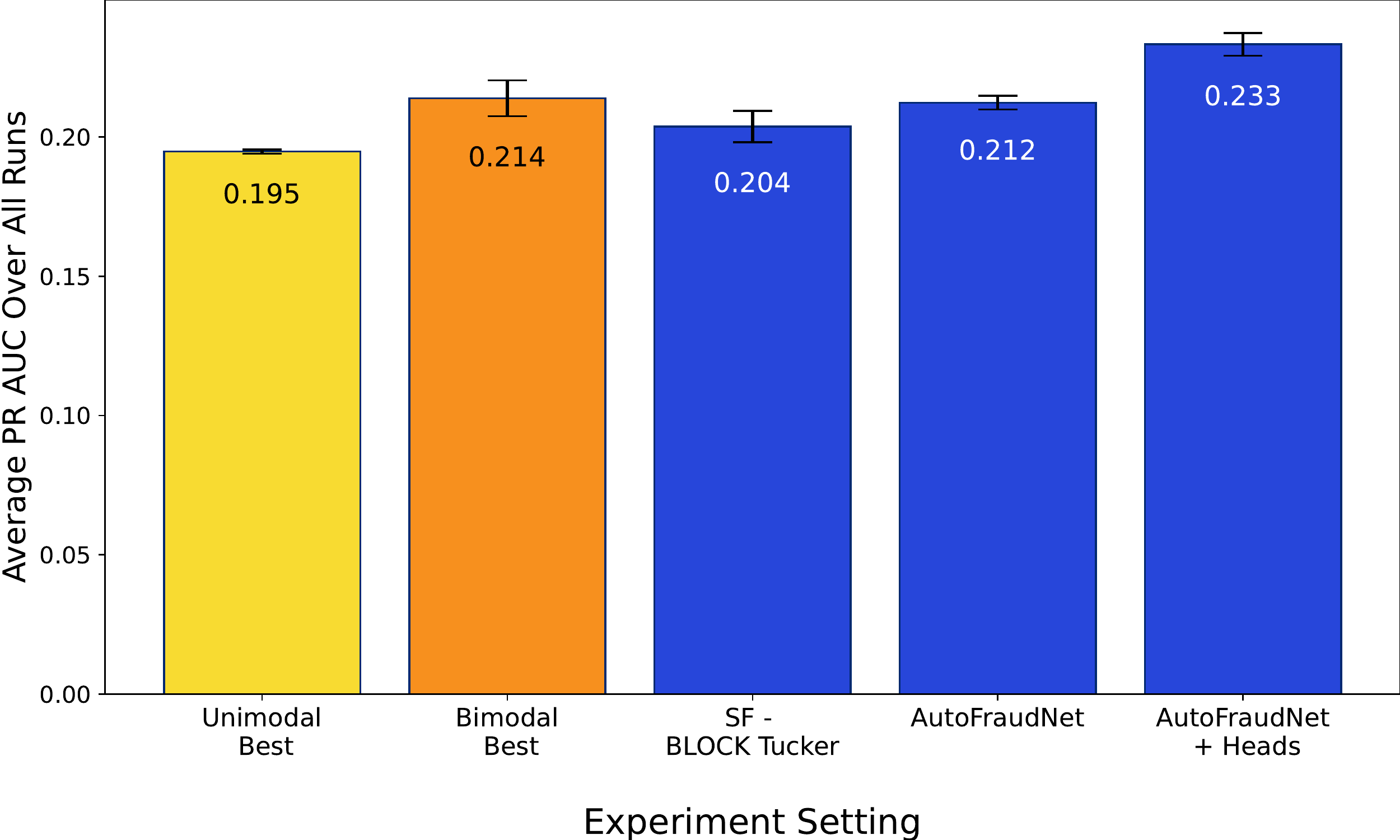}
  \captionof{figure}{Comparison between Unimodal (yellow), Bimodal (orange), and Multimodal (blue) Settings. The corresponding average PR AUC is shown on each bar.}
  \label{fig:general_trends}
\end{minipage}
\end{figure*}

Figure~\ref{fig:bi_FBs} shows a more holistic comparison of the $7$ fusion strategies. Here, we present the average and standard deviation of PR AUC across all runs of $8$ feature pairs. We observe that a subset of fusion strategies perform better than the naïve Concat MLP. More specifically, BLOCK Tucker obtains the highest performance across all fusion strategies, followed by BLOCK, MLB, and MFB. The same trend is evident in Figures~\ref{subfig:bi_heatmap_avg} and \ref{subfig:bi_heatmap_std} along the horizontal axis.

\subsection{Multimodal}
\label{subsec:res-multimodal}
As mentioned in section~\ref{subsec:exp-multimodal}, we use our findings from the bimodal experiments to guide the search space of the multimodal experiments. Therefore, we use features pairs (CDS, SPUD) and (UD, Struct) with BLOCK Tucker in the first fusion layer of our SF framework due to their superior performance. Additionally, we select MFB, MLB, BLOCK, and BLOCK Tucker as candidates for the second fusion layer as they reached better performances than Concat MLP in the bimodal case (section~\ref{subsec:res-bimodal}).

Table~\ref{table:multi_modality} provides a comparison of our proposed framework with the baselines and different multimodal configurations. First, we observe that \textit{Concat MLP - w/o Text} performs better than \textit{Concat MLP - All} in terms of PR AUC. Therefore, we use \textit{Concat MLP - w/o Text} as our baseline. 
Furthermore, we note that BLOCK Tucker is the best candidate for the second layer of fusion followed closely by BLOCK.
Surprisingly, \textit{SF - MLB} and \textit{SF - MFB} do not perform any better than \textit{Concat MLP - w/o Text}.

AutoFraudNet outperforms all other multimodal configurations across all metrics. AutoFraudNet achieves an increase of 2\% in PR AUC when compared to \textit{Concat MLP - w/o Text} and 0.9\% when compared to \textit{SF - BLOCK Tucker}. We believe this is because of the compactness of AutoFraudNet, especially in the second fusion layer. In particular, AutoFraudNet consists of 21.6M parameters (41.4\% fewer parameters compared to SF - BLOCK Tucker). 

AutoFraudNet + Heads further achieves a boost of 2.1\% in PR AUC over AutoFraudNet.
This boost in performance comes with the addition of only 6.4K parameters to AutoFraudNet. AutoFraudNet + Heads also shows a significant increase of $\sim$ 15\%  in Bal. Acc. when compared to \textit{SF - BLOCK Tucker} and \textit{Concat MLP - w/o Text}. This attests to the effectiveness of the multi-head approach~\cite{szegedy2015going} and the compact yet effective architectural design of AutoFraudNet + Heads.

\begin{table*}[ht]
\centering
\caption{Performance comparison of various multimodal configurations}
\label{table:multi_modality}
\begin{tabular}{rcccccccc} 
\toprule
\multirow{2}{*}{Configuration} & \multirow{2}{*}{PR AUC} & \multirow{2}{*}{Bal. Acc.} & \multicolumn{3}{c}{Fraudulent} & \multicolumn{3}{c}{Not Fraudulent}\\
\cmidrule(lr){4-6} \cmidrule(l){7-9}
& & & \mc{P} & \mc{R} & \mc{F1} & \mc{P} & \mc{R} & \mc{F1}\\
\midrule
Concat MLP - All                 & 0.179 & 0.597 & 0.057 & 0.926 & 0.106 & 0.395 & 0.269 & 0.320\\
Concat MLP - w/o Text                & 0.192 & 0.601 & 0.059 & 0.924 & 0.109 & 0.395 & 0.277 & 0.326\\
\midrule
SF - MFB                             & 0.158 & 0.549 & 0.047 & 0.962 & 0.089 & 0.197 & 0.136 & 0.161\\
SF - MLB                             & 0.165 & 0.648 & 0.068 & 0.889 & 0.125 & 0.593 & 0.407 & 0.483\\
SF - BLOCK                           & 0.201 & 0.548 & 0.046 & 0.963 & 0.088 & 0.197 & 0.133 & 0.159 \\
SF - BLOCK Tucker                    & 0.203 & 0.595 & 0.056 & 0.924 & 0.105 & 0.395 & 0.266 & 0.318 \\ \midrule
AutoFraudNet                             & 0.212 & 0.650 & 0.070 & 0.886 & 0.128 & 0.593 & 0.415 & 0.488 \\
AutoFraudNet + Heads                     & \textbf{0.233} & \textbf{0.751} & \textbf{0.092} & \textbf{0.811} & \textbf{0.165} & \textbf{0.989} & \textbf{0.690} & \textbf{0.813} \\ 
\bottomrule
\end{tabular}
\end{table*}

\subsection{Insights}
\label{subsec:res-insights}
In this section we present our overall findings based on the results discussed in the previous sections. Figure~\ref{fig:general_trends} shows the PR AUC for the best performing unimodal (yellow), bimodal (orange), and multimodal (blue) configurations. The PR AUC improves as we move from \textit{Unimodal Best} to  \textit{Bimodal Best} and \textit{Bimodal Best} to \emph{\textit{AutoFraudNet + Heads}}.
This upward trend shows the complementary nature of different sources of information and the benefits of utilizing more data modalities. However, as the high variability between the multimodal configurations suggests, choosing the appropriate fusion strategy is crucial for achieving these benefits.
In fact, choosing the wrong fusion mechanism can even deteriorate performance as we see when comparing \textit{Bimodal Best} with \textit{SF - BLOCK Tucker}.

Our findings suggest that SF frameworks, if carefully designed, can address some of the natural challenges that arise with multimodal learning such as different levels of feature granularity. However, this requires a mindful consideration of data availability and model complexity. The fact that \emph{AutoFraudNet + Heads} outperforms the other SF frameworks despite having considerably fewer parameters highlights the challenges of overfitting and the importance of compactness when designing SF frameworks. 


\section{Limitations \& Future work}
\label{sec:limitations_and_future_work}
Throughout this work, we identified certain limitations that are essential to address and can open up promising paths for future research.

First, to extract textual features from Japanese comments, we fine-tuned BERT, originally pre-trained on an English corpus. But given the differences between the two languages and their underlying grammatical structures, our fine-tuning regime gave very limited success.
On the other hand, our dataset did not have sufficient volumes of Japanese comments to justify large scale pre-training of BERT. Additionally, there are not a lot of pre-trained models for processing Japanese text. As a result, possible future directions entail collection of more claims, and exploration of language models that are pre-trained on relevant Japanese corpuses.

Second, our framework asserts the availability of all modalities by dropping claims with missing features. However, it is quite possible for claims to not have all modalities present in them. A possible next step can involve employing a learned gated function to regulate feature propagation based on their availability. Adding this capability can increase our framework's robustness.

Finally, our framework lacks explainability in its current form. To ensure swift model adoption amongst our end-users, it is imperative to showcase why our framework tags a claim as fraudulent. To this end, a future line of work can include the adoption of attention maps to highlight which parts of an image, or which words/phrases of textual comments, triggered an alarm. Having such abilities to explain model decisions will not only assist performance improvements via error analysis, but also help establish trust in model predictions amongst our end-users.

\section{Conclusion}
\label{sec:conclusions}
In this paper we presented a multimodal reasoning framework, \emph{AutoFraudNet}, to detect fraudulent auto-insurance claims.
To overcome the inherent challenges of multimodal learning, our framework employs a slow fusion paradigm, state-of-the-art fusion blocks, and a compact architectural design.
Our framework utilizes different data modalities including visual data (images) and structural (tabular) metadata.
Through extensive experiments conducted on a real-world dataset we compared our framework to various uni-, bi-, and multi-modal settings and demonstrated its superior performance.
Cognizant of the importance of multimodal learning, especially in fraud detection, we set ourselves to address and improve upon the limitations of \emph{AutoFraudNet}. We hope that our work opens up new avenues of exploration for future research in this direction.

\section*{Acknowledgments}
We would like to thank Laurent Decamp, Mohan Mahadevan, Kush Madlani, Frederick Hoffman and Mathieu Orhan from Tractable, and Parinaz Sobhani, Kyryl Truskovsky, Angeline Yasodhara and Zilun Peng from Georgian for their useful insights.

{\small
\bibliography{aaai23}
}


\end{document}